\documentclass{article}

\usepackage{arxiv}

\usepackage[utf8]{inputenc} 
\usepackage[T1]{fontenc}    
\usepackage[colorlinks=true, citecolor=blue, linkcolor=blue, urlcolor=blue]{hyperref}
\usepackage{url}            
\usepackage{booktabs}       
\usepackage{amsmath}
\usepackage{nicefrac}       
\usepackage{microtype}      
\usepackage{cleveref}       
\usepackage{lipsum}         
\usepackage[dvipsnames, table]{xcolor}
\usepackage{xcolor}
\usepackage{float}
\usepackage{graphicx}
\usepackage{tabularx}
\usepackage{doi}

\usepackage[backend=biber, style=apa]{biblatex}
\usepackage{color}

\graphicspath{{./figs/}}

\DeclareFieldFormat{citehyperref}{%
	\DeclareFieldAlias{bibhyperref}{default}%
	\bibhyperref{#1}}

\savebibmacro{cite}
\savebibmacro{textcite}

\renewbibmacro*{cite}{%
	\printtext[citehyperref]{\restorebibmacro{cite}\usebibmacro{cite}}}

\renewbibmacro*{textcite}{%
	\printtext[citehyperref]{\restorebibmacro{textcite}\usebibmacro{textcite}}}

\title{Digital Pantheon: Simulating and Auditing Coalition Formation with LLM Agents}

\date{}

\newif\ifuniqueAffiliation
\uniqueAffiliationtrue

\ifuniqueAffiliation 
\author{ 
	\hypersetup{hidelinks}
	\href{https://orcid.org/0000-0002-0141-6775}
	{\includegraphics[scale=0.06]{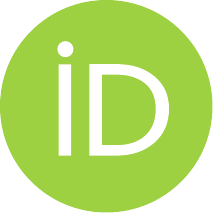}\hspace{1mm}\textcolor{blue}{Dylan Van Mulders}}
	\thanks{Corresponding author} \\
	Ghent University \\
	CVAMO Core Lab \\
	\texttt{dylan.vanmulders@ugent.be} \\
	\And
	\hypersetup{hidelinks}
	\href{https://orcid.org/0000-0002-4502-0764}
	{\includegraphics[scale=0.06]{orcid.pdf}\hspace{1mm}\textcolor{blue}{Matthias Bogaert}} \\
	Ghent University \\
	CVAMO Core Lab \\
	\texttt{matthias.bogaert@ugent.be} \\
	\And
	\hypersetup{hidelinks}
	\href{https://orcid.org/0000-0002-8676-8103}
	{\includegraphics[scale=0.06]{orcid.pdf}\hspace{1mm}\textcolor{blue}{Dirk Van den Poel}} \\
	Ghent University \\
	CVAMO Core Lab \\
	\texttt{dirk.vandenpoel@ugent.be} \\
}

\renewcommand{\undertitle}{ACCEPTED MANUSCRIPT VERSION OF A WORKSHOP PAPER AT ECML-PKDD 2026}

\hypersetup{
pdftitle={Digital Pantheon: Simulating and Auditing \\ Coalition Formation with LLM Agents},
pdfsubject={cs.CL},
pdfauthor={Dylan ~Van Mulders, Matthias ~Bogaert, Dirk ~Van den Poel},
pdfkeywords={Generative Agent-Based Modeling, Agent Based Model, Multi-Agent Negotiation, Large Language Models, AI Alignment and Behavioral Steering, Political Simulation, Explainability.},
}

\begin{document}
\maketitle

\begin{abstract}
	The formation of political coalitions is a complex negotiation driven by both concrete policy objectives and deep-seated ideological convictions. While Large Language Models (LLMs) open new avenues for computational political science, the neutrality and helpfulness biases instilled by Reinforcement Learning from Human Feedback (RLHF) prevent them from sustaining steadfast partisan behaviour. We present a multi-agent framework that reconciles factual grounding with ideological alignment by combining Supervised Fine-Tuning (SFT), Direct Preference Optimization (DPO), and Retrieval-Augmented Generation (RAG): DPO instils aggressive party-specific personas, while a per-party RAG pipeline keeps each agent bounded to its official manifesto. We operationalize the framework on the 2019 Flemish election, deploying the partisan agents in a hub-and-spoke negotiation arbitrated by a formateur. To make the emergent negotiation interpretable, we introduce a Multi-Layered Information Lineage Topology (MILT) that traces every clause in the final agreement back to its manifesto origin and classifies it into five provenance states, a Coalition Influence Score (CIS) that aggregates these traceable contributions to identify which party shaped the agreement, and a real-world grounding pass that benchmarks each simulated provision against the historically adopted coalition agreement. Across three independent simulations the framework yields a stable winner and ranking (N-VA ahead of CD\&V and Open Vld), and manifesto-anchored lineage reliably predicts real-world materialization whereas hallucinated content does not. The result is a transparent, scalable testbed for the ex-ante exploration of party compatibility and formateur-mediated compromise.
\end{abstract}


\section{Introduction}
\label{sec:introduction}

Coalition formation lies at the heart of parliamentary democracy. In multi-party systems such as Belgium's, governing requires protracted negotiation over policy platforms, portfolios, and ideological compromises---shaped not only by rational calculus but by partisan identities and rhetorical traditions that resist formalization, which traditional game-theoretic and agent-based approaches capture structurally but cannot model at the level of language-mediated deliberation.

Large Language Models (LLMs) have recently been used to simulate legislative voting \parencite{li-paa-2024}, draft consensus resolutions \parencite{zhang-policon-2026}, and generate coalition compromises \parencite{briman-2026}, building on a broader LLM-agent paradigm \parencite{wang-2024,xi-2025,sumers-2024} extended to multi-agent societal and policy deliberation \parencite{guo-2024}. Yet, models trained using Reinforcement Learning from Human Feedback (RLHF) are tuned to be neutral and agreeable---antithetical to sustained partisanship: prompt-based personas degrade reasoning \parencite{kim-2024} and lack domain grounding \parencite{jiang-personallm-2024}, while multi-agent debate amplifies conformity and dominant-party bias \parencite{cui-2025,zhang-policon-2026}.

We address this by combining \emph{(i)} Supervised Fine-Tuning (SFT) and Direct Preference Optimization (DPO) to embed party-specific commitments at the parameter level \parencite{agiza-politune-2025}; \emph{(ii)} a Retrieval-Augmented Generation (RAG) pipeline grounding each agent in its party's official manifesto \parencite{wang-2024,li-2025}; and \emph{(iii)} a structured deliberation protocol for autonomous negotiation, informed by work on communication topology \parencite{yang-2026} and decision protocols \parencite{kaesberg-2025,becker-mallm-2025}. We operationalize the framework on Flemish politics after the 2019 election, benchmarking emergent deliberation against the real-world N-VA / CD\&V / Open Vld coalition, and add a dedicated evaluation suite -- a \emph{Multi-Layered Information Lineage Topology} (MILT), a \emph{Coalition Influence Score} (CIS), and a real-world grounding pass against the historically adopted 2019 agreement -- to keep the negotiation interpretable rather than a black box.

Our contributions are: (1) behaviorally steered agents that maintain partisan consistency under extended negotiation; (2) a SFT+DPO+RAG design that is simultaneously ideologically faithful and policy-substantive; (3) MILT and the CIS, an explainability framework attributing every clause in the final agreement to a party and its manifesto origin; (4) an external validation grading the simulated agreement against the historically adopted coalition agreement, showing that manifesto-anchored lineage corresponds to real-world materialization; and (5) a scalable, generative tool for exploring political strategy in democratic coalition formation.

\section{Related Work}
\label{sec:related-work}

Our contribution sits at the intersection of computational \emph{political \& coalition simulation} and \emph{multi-agent debate \& deliberation}. Prior work on LLM-agent architectures -- spanning profiling, memory, planning, action \parencite{wang-2024,xi-2025,sumers-2024,li-2025}, single-to-multi-agent transitions \parencite{guo-2024}, and agent evaluation \parencite{yehudai-2025} -- provides our foundation. However, we highlight, what these systems achieved specifically as \emph{political actors} and where they fall short.

\subsection{Coalition Builders and Political Agents}
The closest line of work models political actors directly. The Political Actor Agent (PAA) \parencite{li-paa-2024} represents individual legislators with scalable profiles and a trustee/delegate/follower planning module, reaching 91.8\% roll-call accuracy and capturing intra-party leader--follower dynamics; yet it forecasts discrete \emph{individual votes} rather than the interactive bargaining through which coalitions form. POLCA \parencite{moghimifar-polca-2024} models multi-party coalition negotiation with LLM agents and benchmarks emergent agreements against real formations, yet evaluates only terminal outcomes, leaving the negotiation a black box with no account of \emph{which} party drove \emph{which} clause. PoliCon \parencite{zhang-policon-2026} contributes a large consensus-drafting benchmark (2{,}225 European Parliament records) scored by an LLM-as-judge, and diagnoses a systematic bias toward dominant-party positions in unsteered models -- the failure mode that undermines faithful partisan simulation -- though its judge-based adjudication is itself opaque. \textcite{briman-2026} formalize coalition compromise as mediation in a semantic document space, guaranteeing outcomes that improve on the status quo, but abstract away from ideology and manifesto language, so the compromises are mathematically principled yet politically ungrounded. UNBench \parencite{liang-unbench-2026} extends evaluation to real institutional behaviour (UN Security Council voting), but again as discrete outcome prediction. On the alignment side, PoliTune \parencite{agiza-politune-2025} shows ideology can be embedded at the parameter level via LoRA and DPO, shifting models along a left--right axis more coherently than prompting, yet does not instantiate \emph{party-specific} platforms or deploy the aligned models in negotiation. A broader negotiation strand -- the large-scale autonomous negotiation competition of \textcite{vaccaro-2025}, the buyer--seller system AgenticPay \parencite{liu-agenticpay-2026}, and theory-of-mind bargaining in peer-to-peer markets \parencite{krohling-2023} -- establishes that LLM agents negotiate competently, but targets commercial rather than ideological, multi-party political settings.

Two capabilities are thus well established -- high-fidelity political behaviour and benchmarkable negotiation -- while two gaps recur: prior systems either predict outcomes without simulating the deliberation, or simulate deliberation without attributing its results to verifiable ideological origins.

\subsection{Multi-Agent Debate and Deliberation}
A second literature studies how multiple LLM agents argue toward a decision. Foundational results show structured debate improves factuality and reasoning \parencite{du-2024} and counters the single-agent ``degeneration of thought'' by encouraging divergent positions \parencite{liang-2024}. Subsequent work refines the protocol: Exchange-of-Thought \parencite{yin-EoT-2023} formalizes communication topologies, voting- versus consensus-based decision rules trade off across task types \parencite{kaesberg-2025,choi-2025}, MALLM \parencite{becker-mallm-2025} decouples debate into composable configurations, and AC$^3$ \parencite{yang-2026} scales deliberation by clustering agents and electing representatives. Critically, this literature exposes its own failure modes: \textcite{cui-2025} document \emph{silent agreement}, where agents capitulate to the majority answer, and \textcite{smit-2024} show strong aggregate debate scores can mask brittle internal dynamics. Both reflect the RLHF-induced pull toward conformity and helpfulness that is fatal to sustained partisanship -- compounded by persona research showing prompt-based role-play degrades reasoning \parencite{kim-2024} and lacks factual grounding \parencite{jiang-personallm-2024}. Argumentation-theoretic work \parencite{amgoud-2005} and explainability-oriented frameworks \parencite{bandara-2025} supply principled vocabularies for offers, concessions, and consensus, but assume cooperative agents and lack grounding in real ideological corpora. On evaluation, adjudication still leans on opaque LLM-judge committees \parencite{zhao-autoarena-2025,becker-mallm-2025}, and \textcite{gritta-2026} argue that outcome accuracy can mask faulty \emph{process}, noting that no existing benchmark jointly covers process transparency and real-world grounding.

\subsection{Positioning our Contribution}
Taken together, prior work leaves four gaps. \emph{(1) Unsteered neutrality and conformity:} debate agents and benchmarked models drift toward consensus and dominant-party positions \parencite{cui-2025,zhang-policon-2026,smit-2024}, and prompt-based personas are too fragile to hold a line \parencite{kim-2024,jiang-personallm-2024}. \emph{(2) Generic, ungrounded ideology:} where ideology is steered it follows a coarse left--right axis \parencite{agiza-politune-2025}, and where coalitions are simulated the policy content is abstract rather than manifesto-grounded \parencite{briman-2026}. \emph{(3) Outcome-only, black-box deliberation:} political simulators predict or score terminal outcomes \parencite{li-paa-2024,moghimifar-polca-2024,liang-unbench-2026} without attributing provisions to their origins. \emph{(4) Opaque adjudication and unmeasured process:} winners and quality are read off LLM-judge preferences \parencite{zhao-autoarena-2025} rather than an auditable account of how agreement was reached \parencite{gritta-2026}.

Our framework targets each gap directly. Against (1) and (2), we abandon prompting for two-stage SFT+DPO alignment on \emph{party-specific} 2019 Flemish manifestos, coupled with a per-party RAG pipeline that keeps each agent factually bounded to its own platform---yielding agents that are neither neutrally helpful nor arbitrarily opinionated but faithfully partisan. Against (3), we stage a structured hub-and-spoke negotiation arbitrated by a formateur, so a coalition agreement \emph{emerges} from deliberation rather than being predicted. Against (3) and (4) jointly, our evaluation suite makes the process auditable: MILT traces every clause of the final agreement back through agents' opening standpoints to its originating manifesto chunk and labels its provenance across five states; the CIS converts these traceable, hallucination-discounting labels into attributable negotiating power; and a real-world grounding layer ($L_{Real}$) benchmarks each provision against the historically adopted agreement -- unifying, for genuinely partisan and manifesto-grounded agents, the process transparency and external fidelity that \textcite{gritta-2026} identify as jointly absent.

\section{Methodology}
\label{sec:methodology}

\begin{figure}[h]
	\centering
	\includegraphics[width=0.85\textwidth]{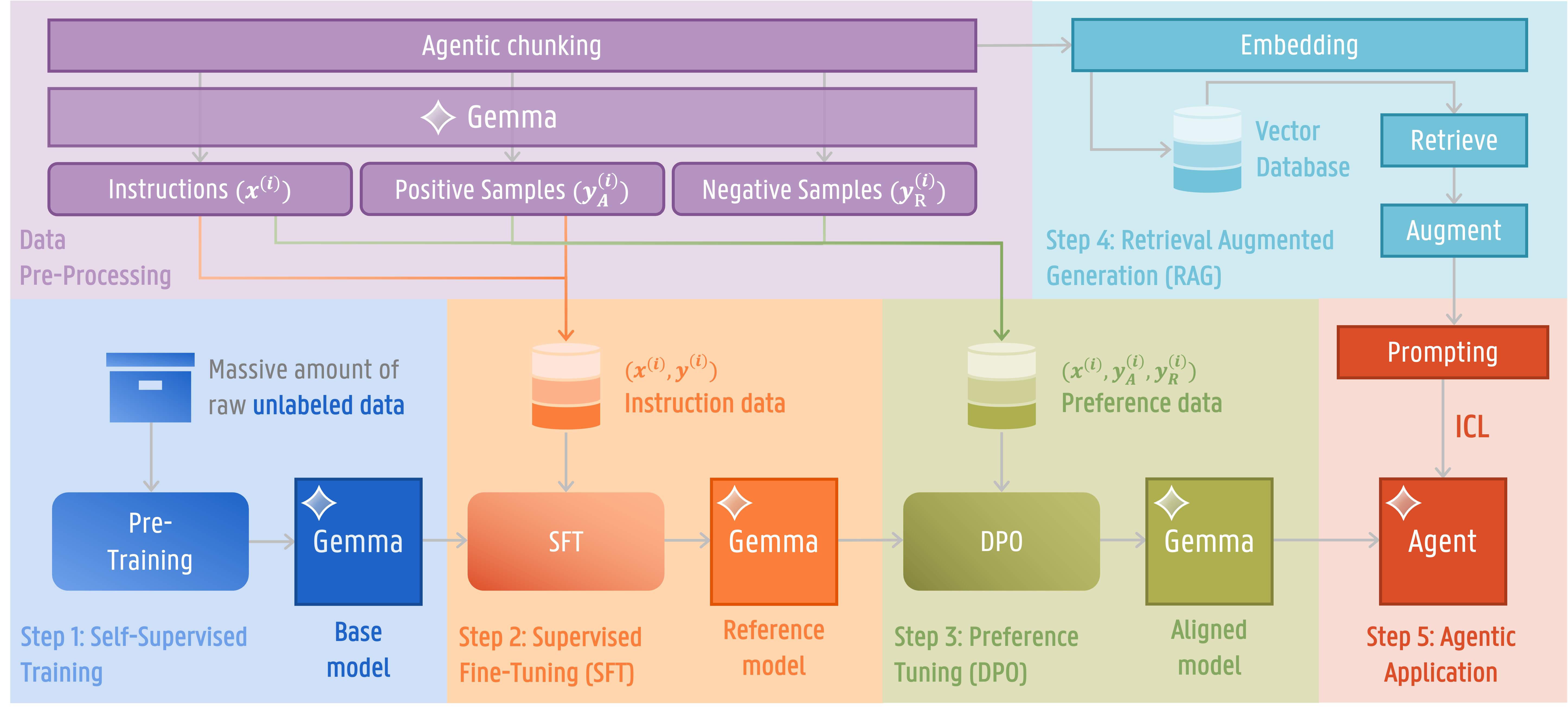}
	\caption{Overview of the individual party alignment model process.}
	\label{fig:methodology}
\end{figure}

We simulate coalition formation with a three-stage pipeline: (1) an isolated RAG knowledge base, (2) a two-stage ideological alignment protocol (using SFT and DPO), and (3) a multi-agent negotiation arena. This decouples factual policy retrieval from persona injection, letting the base Large Language Model (LLM) argue subjectively while staying grounded in official manifestos (see Figure~\ref{fig:methodology}).

\subsection{Data and Document Chunking}
\begin{figure}[h]
	\centering
	\includegraphics[width=0.75\textwidth]{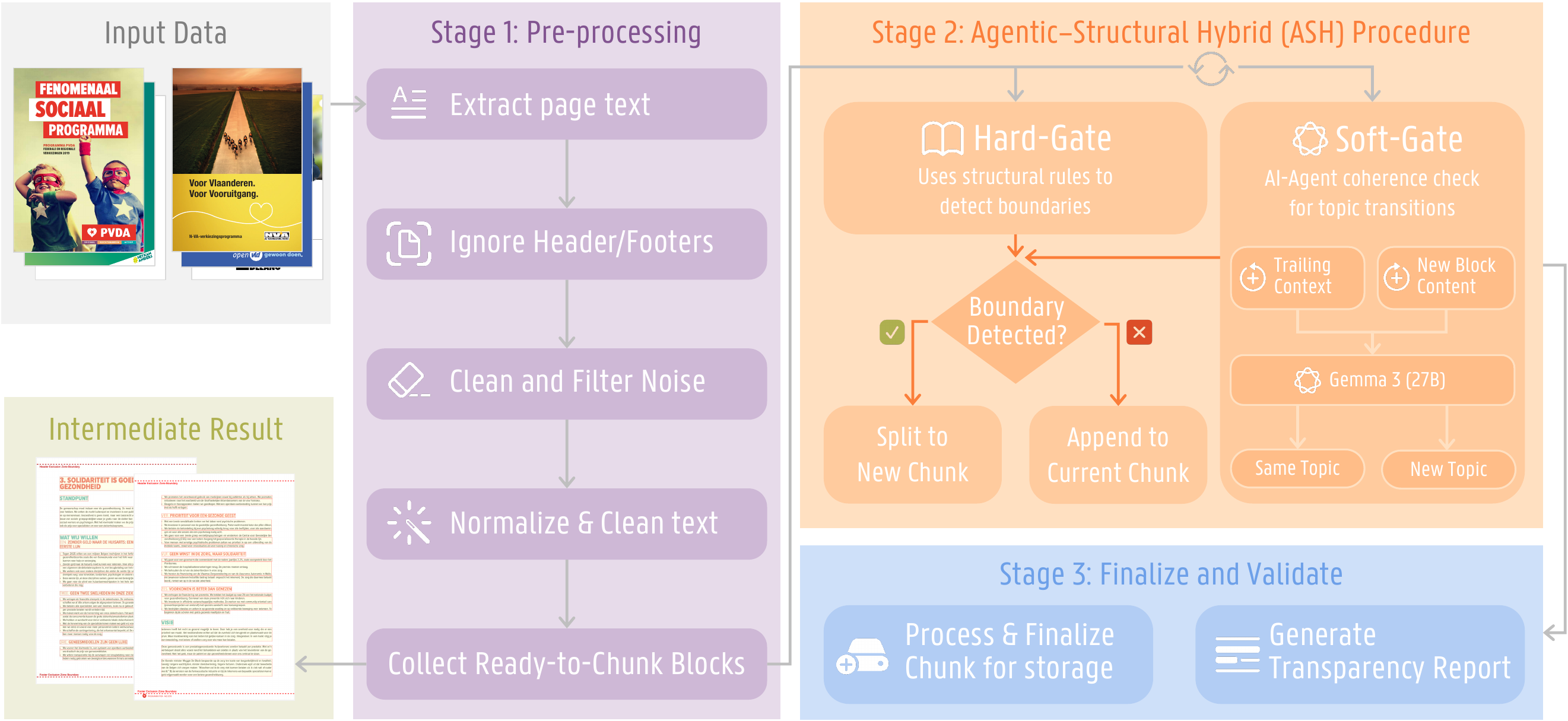}
	\caption{Overview of the hybrid chunking strategy.}
	\label{fig:ASH-overview}
\end{figure}

The corpus comprises the official 2019 electoral manifestos of the major Flemish parties. We partition each with an \emph{Agentic--Structural Hybrid (ASH)} procedure (Figure~\ref{fig:ASH-overview}), a \emph{dual-constraint partitioning problem} where chunk boundaries respect both document hierarchy and local semantic coherence. PDFs are parsed with PyMuPDF \parencite{PyMuPDF-2024} at the dictionary level for per-span bounding boxes and font sizes; margin and sub-minimum-font blocks are discarded, removing headers, page numbers, and footnotes without a layout model. Text is NFKC-normalized with an explicit ligature, custom-font, and smart-quote handling, and intra-paragraph breaks collapsed. Two cascaded constraints follow. The \emph{Hard-Gate} is deterministic: regex detects formal section openings -- numeric subsections (\texttt{1.1}, \texttt{1.1.1}), Article/Section/Chapter markers, capitalized colon-terminated headers -- and forces a split, compensating for LLMs' \emph{structural blindness} and guaranteeing distinct subsections are never conflated even when lexically near-duplicate. The \emph{Soft-Gate} fires only on residual transitions: Gemma3 (27B) \parencite{gemma3-2025} via Ollama \parencite{Ollama-2023}, queried with a structured-output schema at temperature 0.1 on the trailing 800 characters plus the candidate next block, returns a binary same-topic judgement. This hybrid design -- deterministic logic where unambiguous, probabilistic reasoning for fluid transitions -- mitigates the \emph{agentic cost bottleneck} of LLM-driven chunking. Each chunk is persisted immediately, with a visualization PDF for auditing.

\subsection{Preference-pair generation}
The persisted chunks populate both the retrieval index and the tuning datasets. Each chunk over fifteen tokens is passed to Gemma3 (27B) \parencite{gemma3-2025} in a single structured-output call (temperature 0.7) whose JSON schema enforces a list of \textit{(Prompt, Chosen, Rejected)} triplets. Because ASH chunks are thematically coherent but not atomic, the instruction directs the model to enumerate every distinct policy resolution and emit at least one Dutch triplet per chunk through prompting: the \textit{Prompt} is a realistic citizen question leading to that resolution; the \textit{Chosen} response rewrites the resolution in \emph{assistant register} -- the first-person conversational style of an instruction-tuned assistant, versus the manifesto's third-person programmatic prose; the \textit{Rejected} counterfactual is a well-written answer to the same prompt that argues from an opposing political ideology or offers a sterile, non-committal stance. Triplets stream to JSONL in HuggingFace's preference-tuning chat format, so one artefact supports both SFT (\textit{Prompt} $\rightarrow$ \textit{Chosen}) and DPO (the full triplet). ASH's structural integrity gives each resolution unambiguous provenance, making preference pairs auditable and retrievals traceable to a document-tree node.

\subsection{Two-Stage Ideological Alignment (SFT and DPO)}
Standard LLMs' neutrality biases preclude authentic political simulation. We use Gemma-3 (27B) \parencite{gemma3-2025}, quantized to 4-bit via QLoRA on Apple MLX \parencite{apple-mlx-2023}, aligned in two stages. 

\textbf{Stage 1 (SFT)} fine-tunes on the \textit{Chosen} responses to adapt conversational structure and style, establishing what we term the ``Method Actor'' persona: rather than describing the party as an outside observer (\textit{``the party proposes\ldots''}), the model speaks \emph{as} the party (\textit{``we will\ldots''}), defending its manifesto commitments as its own.

\textbf{Stage 2 (DPO)} addresses what SFT cannot: cross-entropy SFT only raises the likelihood of the target text, layering a stylistic overlay on a base model whose ingrained prior for encyclopedic neutrality remains intact, so under adversarial prompting the model reverts to that dominant prior. DPO's contrastive objective instead maximizes the implicit reward gap between \textit{Chosen} and \textit{Rejected} \parencite{rafailov-dpo-2024}, pushing probability mass away from the neutral or opposing responses -- penalizing encyclopedic neutrality and rewarding combative partisan rhetoric (implicit KL-regularization strength $\beta = 0.1$ against the SFT reference policy).

To equalize alignment depth across manifestos of varying length, each party's adapter steps are scaled as $I=\frac{N}{B} \times E$ with $I$ the total number of iterations, $N$ the dataset size, $B$ the batch size, and $E$ the target epochs. SFT uses $B=2$ and $E=5$; DPO uses $B=1$ (given the memory overhead of preference tuning a 27B model) and $E=3$, empirically chosen to prevent catastrophic forgetting. In the DPO stage a low learning rate of $5\times10^{-6}$ resumes and updates the SFT adapter weights.

\subsection{Policy Grounding through RAG}
At inference we supplement the fine-tuned model with RAG: where SFT and DPO calibrate the persona, RAG supplies factual constraints against the ``knowledge boundary problem'' of relying on generalized internal knowledge. The retrieval corpus is the same policy data used in fine-tuning, so arguments derive strictly from the party's positions. Concretely, an isolated ChromaDB \parencite{chroma-2023} pipeline assigns each party its own vector collection in a local SQLite-backed store to prevent cross-contamination. Dense vectors are computed on CPU via paraphrase-multilingual-MiniLM-L12-v2 served by FastEmbed \parencite{fastembed-2023}, giving each agent perfect recall of its manifesto without memory bloat.

\subsection{Multi-Agent Negotiation Arena and Coalition Simulation}
The arena follows a \emph{hub-and-spoke} topology: party agents are spokes, a central \textit{formateur} the hub. The formateur is instantiated from the same fine-tuned LLM as the largest coalition party but is system-prompted to adopt the persona of a neutral, realistic broker; after parties present standpoints it drafts an interim agreement refined across four rounds into the final agreement. This orchestration, the \textit{Digital Pantheon}, turns static QA models into autonomous interacting entities. The four-round horizon is a design choice rather than a measured convergence criterion: in preliminary runs it was the point at which each party's priorities and common ground were fully surfaced, while further rounds added little new negotiation content. For efficiency, adapter hot-swapping keeps the base Gemma-3 weights static in unified memory while party-specific DPO adapters are loaded per turn, simulating all major parties on consumer hardware.

The protocol is an iterative multi-turn debate. Each turn, an active agent processes the dialogue state and runs a semantic search against its policy vector collection; the top-$k$ chunks (cosine similarity $<0.51$, maximum $k=5$) are injected as strict ideological boundaries, so agents cannot concede core promises or hallucinate out-of-line compromises. Under a zero-shot consensus objective, agents articulate non-negotiable \textit{``red lines''}, identify shared objectives, and propose syntheses. Mirroring convention, the largest party assumes the \textit{formateur} role, drafting the final agreement and judging whether balanced consensus is reached. To enforce realistic power dynamics it is given the Banzhaf power indices of all parties \parencite{penrose-1946}. In our initial tests, a non-fine-tuned base formateur favoured moderate parties even with these indices, whereas a party-tuned formateur preserves the leverage they dictate. Logging all rounds -- retrieved chunks, interim drafts, final concessions -- yields a computational model of expected coalition structures. We apply this approach to all 22 policy topics of the Flemish government's portfolio, derived from the official 2019 coalition agreement (Table~\ref{tab:dossier-grounding}).

\subsection{Explainability}
To counter the black-box nature of generative agents and mitigate \textit{prompt artifact bias}, we introduce a \textit{Multi-Layered Information Lineage Topology} (MILT), modelling the negotiation as a temporal directional graph. Rather than parsing voluminous transcripts, we infer the negotiation dynamics ($L_2$) by tracing every policy measure in the final agreement ($L_3$) backward through agents' initial standpoints ($L_1$) to the retrieved manifesto chunks ($L_0$). We deploy Qwen3.6 (27B) via Ollama as a backward Natural Language Inference (NLI) classifier \parencite{qwen3.6-27b,Ollama-2023}: in one JSON inference call it ingests $L_0$, $L_1$, and $L_3$ together, capturing not just \textit{whether} an idea mutated but \textit{when}, distinguishing preemptive dilution ($L_0 \rightarrow L_1$) from reactive concession ($L_1 \rightarrow L_3$). It categorizes provenance into five topological states, separating valid negotiation dynamics from systemic errors:
\begin{enumerate}
	\item \textbf{Direct Lineage (Ideological Preservation):} maps to a party's $L_0$ manifesto and survives $L_1$, the negotiation, and remains intact to $L_3$.
	\item \textbf{Diluted Lineage (Concession):} originates in an $L_0$ manifesto but is measurably weakened, preemptively ($L_0 \rightarrow L_1$) or reactively ($L_1 \rightarrow L_3$).
	\item \textbf{Synthesized Lineage (Logrolling):} combines distinct, competing $L_0$ chunks into a novel hybrid generated through interaction.
	\item \textbf{Pipeline Artifact (Initialization Bias):} present in $L_1$ and $L_3$ but absent from $L_0$, exposing prompt-induced setup noise.
	\item \textbf{Orphan (Debate Hallucination):} appears in $L_3$ but lacks $L_0$/$L_1$ grounding, a spontaneous negotiation-phase hallucination.
\end{enumerate}
The resulting JSON records the policy text, originating party, $L_0$/$L_1$ grounding indicators, topology class, and a qualitative audit trail. By isolating ideological retention from system noise, MILT quantifies negotiation dynamics from the survival of human-authored source material without annotating debate logs.

\subsection{Evaluation: A Coalition Influence Score}
To name the winner from the topology, we map +3 (Direct), +2 (Diluted), +1 (Synthesized) into a \textbf{Coalition Influence Score (CIS)}. Let $P$ denote the set of policy measures in the final agreement, $\Pi$ the set of coalition parties (here $\{$N-VA, CD\&V, Open Vld$\}$), and $T$ the five MILT states; $\pi : P \rightarrow \Pi$ maps each policy to its originating party and $\tau : P \rightarrow T$ to its topology state. For each policy $p$:
\begin{equation}
	\mathrm{CIS}(\pi) \;=\; \sum_{p \,:\, \pi(p) = \pi} w(\tau(p)), \qquad w =
	\begin{cases}
		3 & \tau = \text{Direct } (\tau_{1}) \\
		2 & \tau = \text{Diluted } (\tau_{2})\\
		1 & \tau = \text{Synthesized } (\tau_{3}) \\
		0 & \text{otherwise } (\tau_{4}/ \tau_{5})
	\end{cases}
	\label{eq:cis}
\end{equation}
The gradient captures degrees of ideological victory: $+3$ for undiluted dominance, $+2$ for agenda-setting despite magnitude concessions, $+1$ for anchoring core principles within hybrid compromises. The $3/2/1$ values are an ordinal encoding of ideological survival rather than calibrated magnitudes. Any monotonically decreasing weighting over $\{$\textit{Direct}, \textit{Diluted}, \textit{Synthesized}$\}$ preserves the intended ordering of per-clause influence. Zero-weighting \textit{Pipeline} and \textit{Orphan} prevents hallucinations from inflating scores: novel provisions from parametric prior knowledge lack a causal link to the platforms, and excluding untraceable content is essential for objectively allocating influence.

\subsection{Empirical Validation via Real-World Grounding}
To test external validity we add a \textit{Real-World Grounding} layer ($L_{Real}$) answering the study's ultimate question: did the simulation invent plausible compromises, or recreate those that occurred? Using the same Qwen3.6 (27B) NLI pipeline \parencite{qwen3.6-27b}, we grade every $L_3$ measure -- keeping its MILT class -- against the historically adopted agreement \parencite{regeerakkoord-2019} as \textit{Present} (exact or functional equivalent), \textit{Partially Present} (altered constraints, timelines, or scope), or \textit{Absent}. Unlike the CIS, this comparison grades every provision regardless of class: including the hallucination classes keeps the test fair rather than circular and lets the data reveal whether traceable lineage predicts materialization. That untraceable classes largely fail to appear is an empirical finding, not an artefact, and does not affect the influence scores. Cross-tabulating internal MILT states against external presence then tests whether anchored outcomes (\textit{Direct}, \textit{Diluted}, \textit{Synthesized}) align with reality while systemic errors (\textit{Pipeline}, \textit{Orphan}) fail to materialize.

\section{Results and Discussion}
\label{sec:results-discussion}

A single pipeline execution negotiates the coalition scenario (N-VA, CD\&V, Open Vld) across the 22 policy dossiers exhausting the Flemish government's competences, producing on average 876 classified provenance points over its $L_0$/$L_1$/$L_3$ corpus. To show results reflecting the simulator rather than one stochastic draw, we repeat the simulation three times ($N = 2{,}629$ points), reporting quantities as across-run mean $\pm$ SD; each run is evaluated twice, isolating evaluator noise from run-to-run variation. Every $L_3$ point carries a MILT label, a human-readable audit trail, and a grounding grade against the 2019 coalition agreement (\textit{Present}, \textit{Partial}, \textit{Absent}), summarized per dossier by the realization index $G = (n_{present} + 0.5\,n_{partial}) / N$. This section follows the framework's three pillars: lineage explainability using MILT, CIS to define the winner, and a real-world grounding analysis.

\subsection{Explainability and the transparency of grounding}

Table~\ref{tab:milt-composition} reports each MILT state across the three simulations with mean $\pm$ SD and corpus share. Provenance-bearing provisions (\textit{Direct}, \textit{Diluted}, \textit{Synthesized}) account for nearly all present and most partial judgements, whereas systemic-error classes (\textit{Orphan}, \textit{Pipeline}) are overwhelmingly absent and seldom fully grounded. This ordering is stable across every simulation and pass. Provisions with genuine manifesto provenance are consistently more likely to match the historically adopted 2019 agreement than the model's ungrounded inventions -- a relationship quantified in the grounding analysis below.

\begin{table}[ht]
	\centering
	\small
	\caption{MILT topology composition across the three simulations, pooled across runs.}
	\label{tab:milt-composition}
	\begin{tabular}{lccccc}
		\toprule
		\textbf{MILT topology state} & \textbf{Sim 1} & \textbf{Sim 2} & \textbf{Sim 3} & \textbf{Mean $\pm$ SD} & \textbf{Share} \\
		\midrule
		Direct Lineage ($\tau_{1}$) & 390 & 373 & 387 & 383.3 $\pm$ 9.1 & 43.7\% $\pm$ 0.9\% \\
		Diluted Lineage ($\tau_{2}$) & 116 & 126 & 117 & 119.7 $\pm$ 5.5 & 13.7\% $\pm$ 0.8\% \\
		Synthesized Lineage ($\tau_{3}$) & 119 & 130 & 111 & 120.0 $\pm$ 9.5 & 13.7\% $\pm$ 1.2\% \\
		Pipeline Artifact ($\tau_{4}$) & 62 & 62 & 55 & 59.7 $\pm$ 4.0 & 6.8\% $\pm$ 0.4\% \\
		Orphan (Hallucination) ($\tau_{5}$) & 212 & 175 & 194 & 193.7 $\pm$ 18.5 & 22.1\% $\pm$ 1.7\% \\
		\midrule
		Ideological Retention ($\tau_{1} + \tau_{2}$) & 506 & 499 & 504 & 503.0 $\pm$ 3.6 & 57.4\% $\pm$ 1.0\% \\
		Systemic Error ($\tau_{4} + \tau_{5}$) & 274 & 237 & 249 & 253.3 $\pm$ 18.9 & 28.9\% $\pm$ 1.6\% \\
		\midrule
		\textbf{Total} & \textbf{899} & \textbf{866} & \textbf{864} & \textbf{876.3 $\pm$ 19.7} & \textbf{100\%} \\
		\bottomrule
	\end{tabular}
\end{table}

The test is meaningful only because the framework is transparent end-to-end: each point's audit trail names the originating party, records its $L_0$ and $L_1$ match, and justifies the topology and grounding in natural language. For instance, the guarantee against \textit{``forced school mergers''} traces verbatim to an N-VA manifesto breakpoint and survives into the final ($L_3$) agreement, whereas an invented \textit{``1.15\% of GNP for education''} target is flagged as an \textit{Orphan} with no $L_0$/$L_1$ antecedent and is absent from reality. Every aggregate below thus decomposes into inspectable, party-attributed decisions.

\subsection{From Topology to Winner}
The same provenance names a winner via the CIS (Eq.~\ref{eq:cis}): a Direct or Diluted clause assigns its full weight to the originating party, a Synthesized clause one point to each contributor its audit trail names -- capturing only explicitly attributable influence. Computed per evaluation pass and averaged within each simulation, the winner is a property of the simulator rather than of a single stochastic draw. N-VA wins every simulation and the ranking N-VA $>$ CD\&V $>$ Open Vld is identical across runs: averaged, N-VA commands 40.0\% $\pm$ 3.3\% of attributable influence, ahead of CD\&V (31.8\% $\pm$ 1.8\%) and Open Vld (28.2\% $\pm$ 1.8\%), with non-overlapping means. This is more balanced than the historical 2019 intra-coalition seat distribution (70/124 parliamentary seats), where N-VA held 50.0\% (35/70), CD\&V 27.1\% (19/70), and Open Vld 22.9\% (16/70).

\begin{table}[ht]
	\centering
	\small
	\caption{Coalition Influence Score (CIS) by party across the three simulations; per-simulation cells give the mean $\pm$ SD across the simulation's two evaluation passes.}
	\label{tab:cis}
	\begin{tabular}{lccccc}
		\toprule
		\textbf{Party} & \textbf{Sim 1 CIS} & \textbf{Sim 2 CIS} & \textbf{Sim 3 CIS} & \textbf{Mean share $\pm$ SD} & \textbf{Rank} \\
		\midrule
		N-VA & \begin{tabular}[t]{@{}c@{}}379.0 $\pm$ 19.8 \\ (43.7\%)\end{tabular} & \begin{tabular}[t]{@{}c@{}}318.5 $\pm$ 14.8 \\ (37.2\%)\end{tabular} & \begin{tabular}[t]{@{}c@{}}336.0 $\pm$ 2.8 \\ (39.1\%)\end{tabular} & 40.0\% $\pm$ 3.3\% & 1 \\
		\midrule
		CD\&V & \begin{tabular}[t]{@{}c@{}}257.0 $\pm$ 15.6 \\ (29.7\%)\end{tabular} & \begin{tabular}[t]{@{}c@{}}279.5 $\pm$ 1.4 \\ (32.7\%)\end{tabular} & \begin{tabular}[t]{@{}c@{}}283.5 $\pm$ 9.2 \\ (33.0\%)\end{tabular} & 31.8\% $\pm$ 1.8\% & 2 \\
		\midrule
		Open Vld & \begin{tabular}[t]{@{}c@{}}230.5 $\pm$ 2.1 \\ (26.6\%)\end{tabular} & \begin{tabular}[t]{@{}c@{}}258.5 $\pm$ 7.8 \\ (30.2\%)\end{tabular} & \begin{tabular}[t]{@{}c@{}}239.0 $\pm$ 4.2 \\ (27.8\%)\end{tabular} & 28.2\% $\pm$ 1.8\% & 3 \\
		\bottomrule
	\end{tabular}
\end{table}

Per-pass counts exceed distinct clauses because multi-party and synthesized provisions credit each named party. Composition is as informative as magnitude and equally stable: N-VA's lead rests overwhelmingly on \textit{Direct Lineage} (about three-quarters of its score every run), Open Vld draws most from \textit{Synthesized} cross-party compromises in which it is most named, and CD\&V sits between with an even split of retained and conceded weight. Crediting syntheses to every contributor narrows but never reorders the junior partners. The simulation thus reproduces a canonical coalition signature -- the largest party (N-VA, who is also formateur) sets the agenda, the smaller liberal partner (Open Vld) builds consensus, the centrist (CD\&V) sits between -- and N-VA's advantage concentrates in the \textit{Direct} class the grounding analysis finds most likely to materialize, so the winner's influence is also the most durable.

\subsection{Empirical validation via real-world grounding}

The grounding distribution is stable across the three simulations. Averaged, points are present in 9.9 $\pm$ 2.0\%, partial in 35.9 $\pm$ 0.9\%, and absent in 54.2 $\pm$ 1.5\%, with $G = 0.278 \pm 0.017$; the two passes per run differ by at most 0.3 points in present rate, so residual variation is evaluator noise.

\begin{table}[ht]
	\centering
	\small
	\caption{Grounding distribution per simulation, pooled across runs.}
	\label{tab:grounding-per-sim}
	\begin{tabular*}{\textwidth}{@{\extracolsep{\fill}} lccccc @{}}
		\toprule
		\textbf{Simulation} & \textbf{N} & \textbf{Present} & \textbf{Partial} & \textbf{Absent} & \textbf{G} \\
		\midrule
		Simulation 1 & 899 & 10.1\% & 35.0\% & 54.8\% & 0.276 \\
		Simulation 2 & 866 & 11.8\% & 35.7\% & 52.5\% & 0.296 \\
		Simulation 3 & 864 & 7.8\% & 36.9\% & 55.3\% & 0.262 \\
		\midrule
		\textbf{Across runs} & \textbf{2,629} & \textbf{9.9 $\pm$ 2.0} & \textbf{35.9 $\pm$ 0.9} & \textbf{54.2 $\pm$ 1.5} & \textbf{0.278 $\pm$ 0.017} \\
		\bottomrule
	\end{tabular*}
\end{table}

Splitting by MILT class (Table~\ref{tab:milt-grounding}, pooled) confirms internal lineage predicts external materialization. The three provenance classes (71.1\% of classifications) realize at $G = 0.32$--$0.37$, whereas the two error classes (28.9\%) collapse to a combined $G = 0.164$: \textit{Orphan}, the largest error class, is \textit{Absent} in 75.4\% of cases (438 of 581). Notably, \textit{Synthesized} clauses realize marginally better than \textit{Direct} ones ($G = 0.365$ vs.\ $0.333$), so genuine cross-party compromise survives at least as reliably as unilaterally retained provisions.

Reproducibility is weaker for generation volume than grounding quality, and the gap is itself an artefact of hallucination: distinct proposals per dossier varied widely across runs (\textit{Poverty Alleviation} $53$, $28$, $14$; \textit{Economy and Innovation} $33$, $27$, $58$), and these swings concentrate in the ungrounded topologies, with \textit{Orphan} and \textit{Pipeline} together near 30\% (760 of 2,629) and far less predictable than manifesto-anchored material. At non-zero temperature the model enumerates fluctuating unsupported elaborations around a stable ideological core, so we base all claims on proportions, not raw volumes.

\begin{table}[ht]
	\centering
	\small
	\caption{Grounding outcomes by MILT topology class, pooled across runs.}
	\label{tab:milt-grounding}
	\begin{tabular*}{\textwidth}{@{\extracolsep{\fill}} lcccc @{}}
		\toprule
		\textbf{MILT topology state} & \textbf{Present} & \textbf{Partial} & \textbf{Absent} & \textbf{$G$} \\
		\midrule
		Direct Lineage ($\tau_{1}$) & 13.9\% & 38.9\% & 47.2\% & 0.333 \\
		Diluted Lineage ($\tau_{2}$) & 4.5\% & 42.3\% & 53.2\% & 0.256 \\
		Synthesized Lineage ($\tau_{3}$) & 14.2\% & 44.7\% & 41.1\% & 0.365 \\
		Pipeline Artifact ($\tau_{4}$) & 5.6\% & 35.2\% & 59.2\% & 0.232 \\
		Orphan (Hallucination) ($\tau_{5}$) & 4.0\% & 20.7\% & 75.4\% & 0.143 \\
		\midrule
		Ideological Retention ($\tau_{1} + \tau_{2}$) & 11.7\% & 39.7\% & 48.6\% & 0.315 \\
		Systemic Error ($\tau_{4} + \tau_{5}$) & 4.3\% & 24.1\% & 71.6\% & 0.164 \\
		\bottomrule
	\end{tabular*}
\end{table}

On that basis, a substantial fraction of provisions reached a real-world counterpart despite only four rounds: 45.8 $\pm$ 1.5\% were graded \textit{Present} or \textit{Partial} (pooled \textit{Present} 9.9\%, 95\% CI [8.7, 11.0]; \textit{Partial} 35.9\%, [34.0, 37.7]; \textit{Absent} 54.2\%, [52.3, 56.1]). Realization is sharply uneven, from $G = 0.491$ (\textit{Education}) to $G = 0.147$ (\textit{Finance and Budget}). Per-dossier counts are too sparse for a single run, but run-invariant grounding (Table~\ref{tab:grounding-per-sim}) licenses pooling the three runs; Table~\ref{tab:dossier-grounding} reports the pooled estimates, ordered by $G$.

\begin{table}[ht]
	\centering
	\footnotesize
	\setlength{\tabcolsep}{4pt}
	\caption{Per-dossier grounding distribution, pooled across runs and ordered by $G$.}
	\label{tab:dossier-grounding}
	\begin{tabular*}{\textwidth}{@{\extracolsep{\fill}} lccccc @{}}
		\toprule
		\textbf{Policy domain} & \textbf{N} & \textbf{Present} & \textbf{Partial} & \textbf{Absent} & \textbf{$G$} \\
		\midrule
		Education & 112 & 23.2\% & 51.8\% & 25.0\% & 0.491 \\
		Poverty Alleviation & 95 & 17.9\% & 47.4\% & 34.7\% & 0.416 \\
		Foreign Policy and Tourism & 82 & 18.3\% & 43.9\% & 37.8\% & 0.402 \\
		Heritage & 48 & 6.3\% & 66.7\% & 27.1\% & 0.396 \\
		Justice & 108 & 12.0\% & 41.7\% & 46.3\% & 0.329 \\
		Well-being & 131 & 14.5\% & 35.1\% & 50.4\% & 0.321 \\
		Cohesion, Onboarding and Integration & 140 & 14.3\% & 35.7\% & 50.0\% & 0.321 \\
		Economy and Innovation & 118 & 11.9\% & 38.1\% & 50.0\% & 0.309 \\
		Energy and Climate & 176 & 12.5\% & 32.4\% & 55.1\% & 0.287 \\
		Housing & 152 & 7.9\% & 40.8\% & 51.3\% & 0.283 \\
		Environment & 117 & 6.0\% & 39.3\% & 54.7\% & 0.256 \\
		Media and Public Broadcaster (VRT) & 131 & 6.1\% & 38.9\% & 55.0\% & 0.256 \\
		Animal Welfare & 121 & 4.1\% & 42.1\% & 53.7\% & 0.252 \\
		Mobility and Public Works & 159 & 11.9\% & 25.8\% & 62.3\% & 0.248 \\
		Government Operations and Efficiency & 159 & 5.0\% & 37.7\% & 57.2\% & 0.239 \\
		Work and Social Economy & 141 & 9.2\% & 27.7\% & 63.1\% & 0.230 \\
		Agriculture and Marine Fishing & 116 & 6.9\% & 31.0\% & 62.1\% & 0.224 \\
		Local and Urban Governance & 191 & 11.5\% & 20.4\% & 68.1\% & 0.217 \\
		Equal Opportunities & 82 & 4.9\% & 31.7\% & 63.4\% & 0.207 \\
		Culture, Youth, and Sport & 106 & 3.8\% & 32.1\% & 64.2\% & 0.198 \\
		Brussels and Flemish Periphery & 52 & 0.0\% & 36.5\% & 63.5\% & 0.183 \\
		Finance and Budget & 92 & 1.1\% & 27.2\% & 71.7\% & 0.147 \\
		\bottomrule
	\end{tabular*}
\end{table}

Two regularities organize this ranking, both following from the level of abstraction at which a proposal is pitched rather than its subject. First, full presence (\textit{Present}) concentrates in programmatic dossiers -- only \textit{Education} (23.2\%), \textit{Foreign Policy and Tourism} (18.3\%), and \textit{Poverty Alleviation} (17.9\%) approach or exceed 18\% \textit{Present} -- and for the best-realized dossiers the \textit{Partial} state is modal (\textit{Heritage} 66.7\%, \textit{Education} 51.8\%), indicating thematic correspondence over literal adoption. Second, the lowest-realized dossiers are saturated with quantified fiscal commitments or competences split with the federal level: \textit{Brussels and Flemish Periphery }records no fully grounded proposal and \textit{Finance and Budget} almost none (1.1\%), the former partly because the theme occupies only a brief manifesto passage ($N = 52$). The dominant \textit{Absent} category is thus chiefly fabricated operational detail -- invented budgets, percentages, deadlines, institutions -- not failed thematic reasoning. The \textit{Partial} state localizes this: most partials identify the correct commitment then append a quantified or institutional specification absent from the agreement, while a smaller set encode genuine contradiction (e.g., a proposal to strengthen Unia where the agreement instead winds it down, or one making the social-housing means test voluntary where the agreement makes it compulsory), which our three-state scheme registers as \textit{Absent}.
	
Two implications follow. Lineage topology is an actionable, ex-ante reliability signal: anchored provisions realize far more often than \textit{Orphans} and \textit{Pipeline Artifacts}, so topology can triage output before any real-world referent exists, and the agreement between internal lineage and external grounding makes the two evaluations mutually reinforcing. And because hallucinated operational specificity is both the dominant and a localized failure mode, it is the most addressable -- amenable to constrained decoding or post-hoc verification without altering the thematic reasoning the simulation already performs well.

\section{Conclusion}
\label{sec:conclusion}

We set out to determine not merely which coalition a multi-agent simulation predicts, but how it reaches agreement and which party shapes it. We built a coalition simulator coupling ASH chunking of manifestos, a two-stage protocol that adapts one base model by SFT then steers it into distinct partisan personas via DPO, and a manifesto-grounded retrieval layer keeping each agent bounded to its platform throughout a hub-and-spoke negotiation arbitrated by a tuned formateur. Unlike debate frameworks that assume cooperative agents, our agents are neither neutrally helpful nor arbitrarily opinionated but faithfully partisan -- the property that makes the emergent dynamics worth measuring.

On this simulator we presented the first end-to-end empirical evaluation of \textit{Multi-Layered Information Lineage Topology} (MILT). Over 22 dossiers across three simulations, each evaluated twice ($N = 2{,}629$ points), we showed that (i) the grounding distribution is highly reproducible -- proportions barely move and $G = 0.278 \pm 0.017$ even as individual edge classifications drift; (ii) $57.4\%$ of provisions are anchored in manifestos (\textit{Direct} or \textit{Diluted}), a further $13.7\%$ arise as genuine synthesis, against a $28.9\%$ systemic-error rate dominated by debate-phase hallucination rather than initialization bias; and (iii) the topology decomposes into per-dossier regimes tracking each area's statutory and substantive structure. Validating these labels against the historical 2019 Flemish agreement, internal provenance and real-world materialization reinforce one another: provisions with verifiable lineage realize far more often than orphaned or pipeline content, so topology is an actionable, ex-ante reliability signal even before ground truth exists. Averaged across runs, $45.8 \pm 1.5\%$ of provisions reached a real-world counterpart after only four rounds -- notable external fidelity -- though realization is uneven across dossiers, peaking in \textit{Education}, with the dominant absent category reflecting fabricated operational detail rather than failed thematic reasoning. Building on these classifications, the \textit{Coalition Influence Score} (CIS), a principled $+3/+2/+1$ mapping excluding hallucinated content by construction, identifies N-VA as the clear winner ($40.0 \pm 3.3\%$ of attributable weight, ahead of CD\&V and Open Vld) -- robust across all runs and passes, consistent with N-VA being both largest party and formateur, and concentrated in the \textit{Direct} class most likely to materialize. MILT thus delivers what an end-to-end classifier cannot: a temporally-decomposed, reproducibility-quantified, externally-grounded account of \textit{how} a simulated coalition reaches agreement, and \textit{who} wins it.

Several caveats and extensions remain. The CIS weights clauses equally regardless of budgetary and political stakes (e.g., \textit{Education} outweighing \textit{Animal Welfare}). The simulation is closed, omitting socio-economic, federal, media-salience, and electorate-feedback channels, while realization is conditioned on the single 2019 referent. Finally, the three-state grounding scheme records contradiction merely as absence, and the automated NLI labels remain transparent only through their audit trails. Future work could enrich the simulation: a two-level-game formulation \parencite{putnam-1988} coupling regional and federal bargaining with a media-salience process, architectural ablations (peer-to-peer, partisan, or multi-formateur committees, and alternative formateur instantiations beyond the largest-party convention adopted here), component ablations quantifying the marginal contribution of SFT, DPO, and RAG against prompt-only or non-tuned baselines, and perturbed inputs such as alternative manifestos (e.g., 2024) or counterfactual coalitions. Cross-system replication would then test whether the narrow attrition band, frame-anchoring dominance, and lineage typology are system-level properties of proportional-representation coalition politics or artefacts of the Flemish segmented party system \parencite{de-winter-2006,deschouwer-2013}. Finally, the evaluation could be hardened by validating a sample of the automated NLI labels against human expert annotators, and by moving the CIS beyond uniform clause weighting, including a sensitivity analysis of its $3/2/1$ gradient under alternative weight vectors.

\printbibliography

@misc{li-paa-2024,
	title={\href{https://arxiv.org/abs/2412.07144}{Political Actor Agent: Simulating Legislative System for Roll Call Votes Prediction with Large Language Models}}, 
	author={Hao Li and Ruoyuan Gong and Hao Jiang},
	year={2024},
	eprint={2412.07144},
	archivePrefix={arXiv},
	primaryClass={cs.AI}
}

@misc{zhang-policon-2026,
	title={\href{https://arxiv.org/abs/2505.19558}{PoliCon: Evaluating LLMs on Achieving Diverse Political Consensus Objectives}}, 
	author={Zhaowei Zhang and Xiaobo Wang and Minghua Yi and others},
	year={2026},
	eprint={2505.19558},
	archivePrefix={arXiv},
	primaryClass={cs.CY}
}

@misc{briman-2026,
	title={\href{https://arxiv.org/abs/2506.06837}{AI-Generated Compromises for Coalition Formation}}, 
	author={Eyal Briman and Ehud Shapiro and Nimrod Talmon},
	year={2026},
	eprint={2506.06837},
	archivePrefix={arXiv},
	primaryClass={cs.MA}
}

@article{wang-2024,
	author  = {Wang, Lei and Ma, Chen and Feng, Xueyang and others},
	title   = {\href{https://doi.org/10.1007/s11704-024-40231-1}{A survey on large language model based autonomous agents}},
	journal = {Frontiers of Computer Science},
	year    = {2024},
	volume  = {18},
	number  = {6},
	pages   = {186345}
}

@article{xi-2025,
	author  = {Xi, Zhiheng and Chen, Wenxiang and Guo, Xin and others},
	title   = {\href{https://doi.org/10.1007/s11432-024-4222-0}{The rise and potential of large language model based agents: a survey}},
	journal = {Science China Information Sciences},
	year    = {2025},
	volume  = {68},
	number  = {2},
	pages   = {121101}
}

@misc{sumers-2024,
	title={\href{https://arxiv.org/abs/2309.02427}{Cognitive Architectures for Language Agents}}, 
	author={Theodore R. Sumers and Shunyu Yao and Karthik Narasimhan and Thomas L. Griffiths},
	year={2024},
	eprint={2309.02427},
	archivePrefix={arXiv},
	primaryClass={cs.AI}
}

@inproceedings{guo-2024, 
	author = {Guo, Taicheng and Chen, Xiuying and Wang, Yaqi and others}, 
	title = {\href{https://doi.org/10.24963/ijcai.2024/890}{Large language model based multi-agents: a survey of progress and challenges}}, 
	year = {2024}, 
	isbn = {978-1-956792-04-1},
	booktitle = {Proceedings of the 33rd IJCAI}, 
	articleno = {890}, 
	numpages = {10}, 
	location = {Jeju, Korea}
}

@misc{kim-2024,
	title={\href{https://arxiv.org/abs/2408.08631}{Persona is a Double-edged Sword: Mitigating the Negative Impact of Role-playing Prompts in Zero-shot Reasoning Tasks}}, 
	author={Junseok Kim and Nakyeong Yang and Kyomin Jung},
	year={2024},
	eprint={2408.08631},
	archivePrefix={arXiv},
	primaryClass={cs.CL}
}

@inproceedings{jiang-personallm-2024,
	title = {\href{https://aclanthology.org/2024.findings-naacl.229/}{{P}ersona{LLM}: Investigating the Ability of Large Language Models to Express Personality Traits}},
	author = {Jiang, Hang  and
	Zhang, Xiajie  and
	Cao, Xubo  and others},
	booktitle = {Findings of the ACL: NAACL 2024},
	year = {2024},
	address = {Mexico City, Mexico},
	publisher = {ACL},
	pages = {3605--3627}
}

@misc{cui-2025,
	title={\href{https://arxiv.org/abs/2509.11035}{Free-MAD: Consensus-Free Multi-Agent Debate}}, 
	author={Yu Cui and Hang Fu and Haibin Zhang and Licheng Wang and Cong Zuo},
	year={2025},
	eprint={2509.11035},
	archivePrefix={arXiv},
	primaryClass={cs.AI}
}

@inproceedings{agiza-politune-2025, 
	author = {Agiza, Ahmed and Mostagir, Mohamed and Reda, Sherief}, 
	title = {\href{https://dl.acm.org/doi/10.5555/3716662.3716663}{PoliTune: Analyzing the Impact of Data Selection and Fine-Tuning on Economic and Political Biases in Large Language Models}}, 
	year = {2025}, 
	publisher = {AAAI Press},
	booktitle = {Proceedings of the 2024 AAAI/ACM Conference on AI, Ethics, and Society}, 
	pages = {2–12}, 
	numpages = {11}, 
	location = {San Jose, California, USA}, 
	series = {AIES '24}
}

@inproceedings{li-2025,
	title = {\href{https://aclanthology.org/2025.coling-main.652/}{A Review of Prominent Paradigms for {LLM}-Based Agents: Tool Use, Planning (Including {RAG}), and Feedback Learning}},
	author = {Li, Xinzhe},
	booktitle = {Proceedings of the 31st International Conference on Computational Linguistics},
	year = {2025},
	address = {Abu Dhabi, UAE},
	publisher = {ACL},
	pages = {9760--9779}
}

@article{yang-2026,
	author  = {Yang, Liancheng and Li, Sining and Deng, Aobo},
	title   = {\href{https://doi.org/10.1007/s11265-025-01978-3}{Dynamic Consensus Communication Mechanism for Large Language Model-Based Multi-Agent Systems}},
	journal = {Journal of Signal Processing Systems},
	year    = {2026},
	volume  = {98},
	number  = {1},
	pages   = {10}
}

@inproceedings{kaesberg-2025,
	title = {\href{https://aclanthology.org/2025.findings-acl.606/}{{Voting or Consensus? Decision-Making in Multi-Agent Debate}}},
	author = {Kaesberg, Lars Benedikt  and Becker, Jonas  and Wahle, Jan Philip  and others},
	booktitle = {Findings of the ACL: ACL 2025},
	year = {2025},
	address = {Vienna, Austria},
	publisher = {ACL},
	pages = {11640--11671},
	ISBN = {979-8-89176-256-5}
}

@inproceedings{becker-mallm-2025,
	title = {\href{https://aclanthology.org/2025.emnlp-demos.29/}{{MALLM}: Multi-Agent Large Language Models Framework}},
	author = {Becker, Jonas  and Kaesberg, Lars Benedikt  and Bauer, Niklas  and others},
	booktitle = {Proceedings of the 2025 Conference on EMNLP: System Demonstrations},
	year = {2025},
	address = {Suzhou, China},
	publisher = {ACL},
	pages = {418--439},
	ISBN = {979-8-89176-334-0}
}

@misc{yehudai-2025,
	title={\href{https://arxiv.org/abs/2503.16416}{Survey on Evaluation of LLM-based Agents}}, 
	author={Asaf Yehudai and Lilach Eden and Alan Li and others},
	year={2025},
	eprint={2503.16416},
	archivePrefix={arXiv},
	primaryClass={cs.AI}
}

@inproceedings{du-2024, 
	author = {Du, Yilun and Li, Shuang and Torralba, Antonio and others}, 
	title = {\href{https://dl.acm.org/doi/10.5555/3692070.3692537}{Improving factuality and reasoning in language models through multiagent debate}}, 
	year = {2024}, 
	publisher = {JMLR.org}, 
	booktitle = {ICML'24: Proceedings of the 41st ICML}, 
	articleno = {467}, 
	numpages = {31}, 
	location = {Vienna, Austria}
}

@inproceedings{yin-EoT-2023,
	title = {\href{https://aclanthology.org/2023.emnlp-main.936/}{Exchange-of-Thought: Enhancing Large Language Model Capabilities through Cross-Model Communication}},
	author = {Yin, Zhangyue  and Sun, Qiushi  and Chang, Cheng  and others},
	booktitle = {Proceedings of the 2023 Conference on EMNLP},
	year = {2023},
	address = {Singapore},
	publisher = {ACL},
	pages = {15135--15153}
}

@misc{bandara-2025,
	title={\href{https://arxiv.org/abs/2512.21699}{Towards Responsible and Explainable AI Agents with Consensus-Driven Reasoning}}, 
	author={Eranga Bandara and Tharaka Hewa and Ross Gore and others},
	year={2025},
	eprint={2512.21699},
	archivePrefix={arXiv},
	primaryClass={cs.AI}
}

@inproceedings{amgoud-2005,
	title={\href{https://dl.acm.org/doi/pdf/10.1145/1082473.1082555}{Towards a formal framework for the search of a consensus between autonomous agents}},
	author={Amgoud, Leila and Belabbes, Sihem and Prade, Henri},
	booktitle={Proceedings of the 4th international joint conference on Autonomous agents and multiagent systems},
	pages={537--543},
	year={2005}
}

@article{putnam-1988,
	ISSN = {00208183, 15315088},
	author = {Robert D. Putnam},
	journal = {International Organization},
	number = {3},
	pages = {427--460},
	publisher = {[The MIT Press, Cambridge University Press, International Organization Foundation]},
	title = {\href{http://www.jstor.org/stable/2706785}{{Diplomacy and Domestic Politics: The Logic of Two-Level Games}}},
	urldate = {2026-05-04},
	volume = {42},
	year = {1988}
}

@software{chroma-2023,
	author = {{Chroma-core}},
	title = {\href{https://github.com/chroma-core/chroma}{Chroma: The AI-native open-source embedding database}},
	version = {0.4.x},
	year = {2023}
}

@software{fastembed-2023,
	author = {{Qdrant Team}},
	title = {\href{https://github.com/qdrant/fastembed}{{FastEmbed}: Fast, light, accurate library built for retrieval embedding generation}},
	version = {0.8.0},
	year = {2023}
}

@article{gemma3-2025,
	title   = {\href{https://arxiv.org/abs/2503.19786}{Gemma 3 Technical Report}},
	author  = {{Gemma Team} and Aishwarya Kamath and Johan Ferret and others},
	year = {2025},
	archivePrefix={arXiv},
	primaryClass={cs.AI}
}

@software{apple-mlx-2023,
	author = {Awni Hannun and Jagrit Digani and Angelos Katharopoulos and Ronan Collobert},
	title = {\href{https://github.com/ml-explore/mlx}{{MLX}: Efficient and flexible machine learning on Apple silicon}},
	version = {0.21.1},
	year = {2023}
}

@article{de-winter-2006,
	author = {De Winter, Lieven and Swyngedouw, Marc and Dumont, Patrick},
	title = {\href{https://doi.org/10.1080/01402380600968836}{Party system(s) and electoral behaviour in Belgium: From stability to balkanisation}},
	journal = {West European Politics},
	volume = {29},
	number = {5},
	pages = {933--956},
	year = {2006},
	publisher = {Routledge}
}

@article{deschouwer-2013,
	author = {Kris Deschouwer and Min Reuchamps},
	title = {\href{https://doi.org/10.1080/13597566.2013.773896}{The Belgian Federation at a Crossroad}},
	journal = {Regional \& Federal Studies},
	volume = {23},
	number = {3},
	pages = {261--270},
	year = {2013},
	publisher = {Routledge}
}

@software{PyMuPDF-2024,
	author = {Jorj McKie and Ruikai Liu},
	title = {\href{{https://github.com/pymupdf/PyMuPDF}}{{PyMuPDF} Documentation}},
	version = {1.24.2},
	year = {2024}
}

@software{Ollama-2023,
	author = {{Ollama Team}},
	title = {\href{{https://github.com/ollama/ollama}}{Ollama Documentation}},
	version = {0.1.48},
	year = {2023}
}

@misc{qwen3.6-27b,
	title  = {\href{{https://qwen.ai/blog?id=qwen3.6-27b}}{{Qwen3.6-27B}: Flagship-Level Coding in a {27B} Dense Model}},
	author = {{Qwen Team}},
	year   = {2026}
}

@article{penrose-1946,
	ISSN = {09528385, 23972335},
		title = {\href{http://www.jstor.org/stable/2981392}{The Elementary Statistics of Majority Voting}},
	author = {L. S. Penrose},
	journal = {Journal of the Royal Statistical Society},
	number = {1},
	pages = {53--57},
	publisher = {[Oxford University Press, Royal Statistical Society]},
	volume = {109},
	year = {1946}
}

@misc{liu-agenticpay-2026,
	title= {\href{https://arxiv.org/abs/2602.06008}{AgenticPay: A Multi-Agent LLM Negotiation System for Buyer-Seller Transactions}}, 
	author={Xianyang Liu and Shangding Gu and Dawn Song},
	year={2026},
	archivePrefix={arXiv},
	primaryClass={cs.AI}
}

@misc{moghimifar-polca-2024,
	title={\href{https://arxiv.org/abs/2402.11712}{Modelling Political Coalition Negotiations Using LLM-based Agents}}, 
	author={Farhad Moghimifar and Yuan-Fang Li and Robert Thomson and Gholamreza Haffari},
	year={2024},
	archivePrefix={arXiv},
	primaryClass={cs.CL}
}

@inproceedings{gritta-2026,
	title = {\href{https://aclanthology.org/2026.findings-eacl.140/}{Process Evaluation for Agentic Systems}},
	author = {Gritta, Milan  and Paul, Debjit  and Li, Xiaoguang  and others},
	booktitle = {Findings of the ACL: {EACL} 2026},
	year = {2026},
	address = {Rabat, Morocco},
	publisher = {ACL},
	pages = {2678--2692},
	ISBN = {979-8-89176-386-9}
}

@article{krohling-2023,
	title = {\href{https://www.sciencedirect.com/science/article/pii/S0952197623000714}{Artificial Theory of Mind in contextual automated negotiations within peer-to-peer markets}},
	author = {Dan E. Kröhling and Omar J.A. Chiotti and Ernesto C. Martínez},
	journal = {Engineering Applications of Artificial Intelligence},
	volume = {120},
	pages = {105887},
	year = {2023},
	issn = {0952-1976}
}

@article{liang-unbench-2026, 
	title={\href{https://ojs.aaai.org/index.php/AAAI/article/view/37040}{Benchmarking LLMs for Political Science: A United Nations Perspective}}, 
	author={Liang, Yueqing and Yang, Liangwei and Wang, Chen and others}, 
	volume={40}, 
	number={1}, 
	journal={Proceedings of the AAAI Conference on Artificial Intelligence}, 
	year={2026},  
	pages={738–745} 
}

@misc{vaccaro-2025,
	title={\href{https://arxiv.org/abs/2503.06416}{Advancing AI Negotiations: A Large-Scale Autonomous Negotiation Competition}}, 
	author={Michelle Vaccaro and Michael Caosun and Harang Ju and Sinan Aral and Jared R. Curhan},
	year={2025},
	archivePrefix={arXiv},
	primaryClass={cs.AI}
}

@inproceedings{zhao-autoarena-2025,
	title = {\href{https://aclanthology.org/2025.acl-long.223/}{Auto-Arena: Automating {LLM} Evaluations with Agent Peer Battles and Committee Discussions}},
	author = {Zhao, Ruochen  and Zhang, Wenxuan  and Chia, Yew Ken  and others},
	booktitle = {Proceedings of the 63rd Annual Meeting of the ACL (Volume 1: Long Papers)},
	year = {2025},
	address = {Vienna, Austria},
	publisher = {ACL},
	pages = {4440--4463},
	ISBN = {979-8-89176-251-0}
}

@inproceedings{smit-2024, 
	author = {Smit, Andries and Grinsztajn, Nathan and Duckworth, Paul and others}, 
	title = {\href{https://dl.acm.org/doi/10.5555/3692070.3693936}{Should we be going MAD? A Look at Multi-Agent Debate Strategies for LLMs}}, 
	year = {2024}, 
	publisher = {JMLR.org}, 
	booktitle = {Proceedings of the 41st International Conference on Machine Learning}, 
	articleno = {1866}, 
	numpages = {23}, 
	location = {Vienna, Austria}, 
	series = {ICML'24} 
}

@inproceedings{liang-2024,
	title = {\href{https://aclanthology.org/2024.emnlp-main.992/}{Encouraging Divergent Thinking in Large Language Models through Multi-Agent Debate}},
	author = {Liang, Tian  and He, Zhiwei  and Jiao, Wenxiang  and others},
	booktitle = {Proceedings of the 2024 Conference on EMNLP},
	month = {nov},
	year = {2024},
	address = {Miami, Florida, USA},
	publisher = {ACL},
	pages = {17889--17904}
}

@inproceedings{choi-2025,
	author = {Choi, Hyeong Kyu and Zhu, Jerry and Li, Sharon},
	booktitle = {Advances in Neural Information Processing Systems},
	pages = {101732--101764},
	publisher = {Curran Associates, Inc.},
	title = {\href{https://proceedings.neurips.cc/paper_files/paper/2025/file/934252acd87f254d5d4672fbde283bd2-Paper-Conference.pdf}{Debate or Vote: Which Yields Better Decisions in Multi-Agent Large Language Models?}},
	volume = {38},
	year = {2025}
}

@misc{regeerakkoord-2019,
	author = {{Vlaamse Regering}},
	title = {\href{https://www.vlaanderen.be/publicaties/regeerakkoord-van-de-vlaamse-regering-2019-2024}{Regeerakkoord van de {Vlaamse} {Regering} 2019--2024 [Coalition Agreement of the {Flemish} Government 2019--2024]}},
	year = {2019},
}

@article{rafailov-dpo-2024,
	author = {Rafailov, Rafael and Sharma, Archit and Mitchell, Eric and others}, 
	title = {\href{https://papers.nips.cc/paper/2023/hash/a85b405ed65c6477a4fe8302b5e06ce7-Abstract-Conference.html}{Direct preference optimization: Your language model is secretly a reward model}}, 
	year = {2024}, 
	journal = {Adv. Neural Inf. Process. Syst.}, 
	volume = {36},
	pages = {53728--53741} 
}

\end{document}